\documentclass[letterpaper]{article}
\usepackage{aaai}

\usepackage{times}
\usepackage{xcolor}
\usepackage{soul}
\usepackage[utf8]{inputenc}
\usepackage[small]{caption}

\usepackage{url}
\usepackage{ifthen}
\usepackage[authoryear]{natbib}
\usepackage{enumitem}

\usepackage{bm}
\usepackage{cancel}
\usepackage{amsmath}
\usepackage{amssymb}
\usepackage{mathtools}
\usepackage{subfigure}
\usepackage[linesnumbered, ruled]{algorithm2e}

\SetKwRepeat{Do}{do}{while}
\SetKwInOut{Parameter}{Parameter}
\DeclarePairedDelimiter\abs{\lvert}{\rvert}

\DeclarePairedDelimiterX{\infdivx}[2]{(}{)}{#1\;\delimsize\|\;#2}

\newcommand{\argmin}{\mathop{\mathrm{argmin}}}
\newcommand{\argmax}{\mathop{\mathrm{argmax}}}

\newcommand{\indep}{\mathop{\perp\!\!\!\perp}}

\title{Structure Learning of Markov Random Fields through Grow-Shrink Maximum Pseudolikelihood Estimation}
\author{
Yuya Takashina$^1$, 
Shuyo Nakatani$^2$, 
Masato Inoue$^1$
\\ 
$^1$ Dept. of Electrical Engineering and Bioscience, Sch. of Advanced Science and Engineering, Waseda University\\
$^2$ Cybozu Labs, Inc., Tokyo, Japan\\
takashina2051@toki.waseda.jp,
nakatani@labs.cybozu.co.jp,
masato.inoue@eb.waseda.ac.jp
}

\begin{document}

\maketitle

\begin{abstract}
Learning the structure of Markov random fields (MRFs) plays an important role in multivariate analysis. The importance has been increasing with the recent rise of statistical relational models since the MRF serves as a building block of these models such as Markov logic networks. There are two fundamental ways to learn structures of MRFs: methods based on parameter learning and those based on independence test. The former methods more or less assume certain forms of distribution, so they potentially perform poorly when the assumption is not satisfied. The latter can learn an MRF structure without a strong distributional assumption, but sometimes it is unclear what objective function is maximized/minimized in these methods. In this paper, we follow the latter, but we explicitly define the optimization problem of MRF structure learning as maximum pseudolikelihood estimation (MPLE) with respect to the edge set. As a result, the proposed solution successfully deals with the {\em symmetricity} in MRFs, whereas such symmetricity is not explicitly taken into account in most existing independence test techniques. The proposed method achieved higher accuracy than previous methods when there were asymmetric dependencies in our experiments.
\end{abstract}

\section{Introduction}
Learning the structure of Markov random fields (MRFs) plays an important role in multivariate analysis \citep{koller2009probabilistic}. The importance has been increasing with the recent rise of statistical relational models since the MRF serves as a building block of these models such as Markov logic networks \citep{richardson2006markov}. If a set of random variables $\bm{X}$ forms an MRF with respect to a graph $G = (V, E)$, it satisfies three Markov properties \citep{rue2005gaussian}:
\begin{itemize}
\item{\em Pairwise Markov:} $X_i \indep X_j \mid \bm{X}_{V\backslash\{i, j\}}$ if $e_{ij} \not\in E$ and $i\neq j$, where $\bm{X}_{V\backslash\{i, j\}}$ denotes the variables except $X_i$ and $X_j$.
\item{\em Local Markov:} $X_i \indep \bm{X}_{V\backslash N_{E}[i]} \mid \bm{X}_{N_{E}(i)}$ for every $i \in V$ where $N_E(i) \equiv \{j \mid j\neq i,~~ e_{ij} \in E ~~\text{or}~~ e_{ji} \in E\}$ is the open neighborhood of $i$, and $N_E[i] \equiv N_E(i) \cup \{i\}$ denotes the closed neighborhood of $i$. 


\item{\em Global Markov:} $X_{A} \indep X_{B}\mid X_{S}$ for all disjoint sets $A, B, S \subset V$ where $S$ separates $A$ and $B$, and $A$ and $B$ are non-empty.
\end{itemize}
In the rest of this paper, we will omit the subscript $E$ in $N_E(i)$ and $N_E[i]$ for brevity when it is obvious from the context which edge set is intended. In structure learning of MRFs, one learns the edge set $E$ from data.

There are two fundamental ways to learn structures of MRFs that utilize two different Markov properties. In algorithms based on {\em parameter estimation}, one assumes a certain form of distribution and learns the underlying graph by estimating the parameters of the distribution, which correspond to the {\em pairwise Markov} property. A basic structure learning algorithm based on parameter estimation assumes that the observations obey a Gaussian distribution with mean $\bm\mu$ and covariance $\bm\Theta^{-1}$, and it estimates $\bm\Theta$, which corresponds to the graph structure, i.e.,
$$
e_{ij} \notin E ~\Leftrightarrow~ \Theta_{ij} = 0.
$$
In the early work, the sparse precision matrix estimation problem was formulated as separate nodewise regressions regarding a variable in an MRF as a target and the other variables as features \citep{meinshausen2006high}. Later, the optimization problem was posed as regularized negative log likelihood minimization \citep{banerjee2008model}, where the graphical lasso \citep{friedman2008sparse} improves its computational cost significantly. Because likelihood optimization remains computationally challenging in many cases, methods using pseudolikelihood \citep{besag1975statistical} were developed in continuous settings \citep{peng2009partial, friedman2010applications}, discrete settings \citep{hofling2009estimation, ravikumar2010high, guo2010joint}, and discrete-continuous mixed settings \citep{lee2015learning}. However, these approaches more or less assume that the random variables obey a specific class of distribution; therefore, the range of application is restricted by the assumptions.

On the other hand, structure learning algorithms based on {\em independence test} utilize the {\em local Markov} property of MRFs. Specifically, these methods learn MRF structure by finding the local neighborhood $N(i)$ for every node. In 1968, Chow and Liu introduced the first systematic study to learn a dependence tree between random variables by finding a maximum spanning tree \citep{chow1968approximating}. Since then, algorithms for more general graphs have been developed. Grow-shrink Markov network (GSMN) \citep{bromberg2009efficient} repeats conditional independence tests under a grow-shrink strategy to identify the existence/absence of edges. Incremental association Markov blanket (IAMB) \citep{tsamardinos2003algorithms} improved the `static' methodology in the grow-shrink algorithm by making it into a `dynamic' one, applying greedy search to neighborhood selection for each variable. Later, a similar greedy solution was proposed with a proof of correctness under specific assumptions \citep{netrapalli2010greedy}. However, in contrast to parameter estimation techniques, it is sometimes unclear what objective function is maximized/minimized in independence-test-based techniques. Moreover, these approaches seem to be unaware of the {\em symmetricity} in MRFs, that is, if a node $i$ is adjacent to another node $j$, then $j$ is adjacent to $i$, too.

In this paper, we follow the independence test techniques and propose a novel structure learning algorithm based on pseudolikelihood maximization. Unlike parameter estimation techniques, we do not assume a certain form of the underlying distribution. Instead, we regard the edge set in graph as a parameter, and we minimize the expectation of negative log pseudolikelihood with respect to the edge set. Since we do not know the concrete form of the distribution, we cannot compute the pseudolikelihood directly from a single observation. Still, we can compute the objective function from observations by taking an expectation over the true distribution and approximating it by the sample mean. The objective function can be greedily minimized, though there is no guarantee of convergence to the optimal value. As a result, the proposed solution successfully deal with the symmetricity in MRFs explicitly. We experimentally evaluated the performance of the proposed method on small synthetic datasets in comparison with several existing algorithms.

\section{Preliminaries}
\subsection{Maximum pseudolikelihood estimation (MPLE)}
Given a set of random variables $\bm{X}$ and a likelihood $p(\bm{x}\mid \mathcal{H}, \Theta)$, where $(\mathcal{H}, \Theta)$ denotes a pair of any distributional assumption and its parameters, the pseudolikelihood \citep{besag1975statistical} is defined as a product of conditional distributions
\begin{equation}
\prod_{i=1}^D p(x_i \mid \bm{x}_{V\backslash \{i\}}, \mathcal{H}, \Theta). \label{eq: pseudo-likelihood}
\end{equation}
For example, if $\mathcal{H}$ is the Gaussian distribution and $\Theta$ is a pair of mean and covariance matrix, then the pseudolikelihood also becomes a Gaussian distribution. Specifically, if $\bm{X}$ forms an MRF with respect to a graph $G = (V, E)$, then the pseudolikelihood is:
$$
\prod_{i=1}^D p(x_i \mid \bm{x}_{N(i)}, \mathcal{H}, \Theta).
$$

In the typical setting of maximum pseudolikelihood estimation (MPLE), negative log pseudolikelihood of observations is minimized with respect to the parameter $\Theta$. Whereas the computation over likelihood is generally intractable because it requires marginalization over a large number of variables, that over the pseudolikelihood can be computed more efficiently.

\subsection{Chow-Liu algorithm}
The Chow-Liu algorithm \citep{chow1968approximating} is a structure learning algorithm for Bayesian networks whose likelihood is defined as:
$$
p(\bm{x}\mid A) \equiv \prod_{i=1}^D p(x_i\mid \bm{x}_{pa(i)}),
$$
where $pa(i)$ denotes the parents of the node $i$ in a directed acyclic graph $G = (V, A)$. Straightforwardly maximizing the joint distribution of samples with respect to arrow (directed edge) set $A$ leads to an optimization problem
\begin{equation}
\max_A \sum_{i\in V} I(X_i; \bm{X}_{pa(i)}), \label{eq: max-mi}
\end{equation}
where $I(X, Y)$ denotes mutual information (MI) between random variables $X$ and $Y$ \citep{campos2006scoring}. When the graph $G$ is a tree, that is, $\abs{pa(i)}$ is at most $1$, then $pa(i)$ can be simply replaced by another node $j$, resulting in the optimization problem
$$
\max_A \sum_{(i, j)\in A} I(X_i; X_j).
$$
This objective function can be greedily maximized. Once the optimal structure is obtained, the direction of each edge will be determined.

\subsection{Grow-shrink Markov network (GSMN)}
Grow-shrink Markov network (GSMN) \citep{bromberg2009efficient} estimates the structure of MRFs by finding the local neighborhood of each variable (also called the Markov blanket of the variable). Given a set of variables $\bm{X} \equiv \{X_i \mid i\in V\}$, where $V$ is the set of indices, the local neighborhood $N(i)$ is the minimum subset of $V$ such that $X_i \indep \bm{X}_{V\backslash N(i)}\mid \bm{X}_{N(i)}$. Specifically, GSMN repeats conditional independence tests to learn the local neighborhood of each node.

There are basically two strategies to conducting those tests: the grow strategy (starting with a null graph and adding edges) and the shrink strategy (starting with a complete graph and removing edges). The grow-shrink algorithm \citep{margaritis2000bayesian}, which was originally intended to learn the structure of Bayesian networks, combines both strategies to avoid the {\em nested effect}, meaning that an edge cannot be removed once it has been added in the grow algorithm, and an edge cannot be added once it has been removed in the shrink algorithm \citep{vergara2014review}. GSMN applied this grow-shrink strategy to learning the structure of MRFs.

GSMN treats the symmetricity of MRFs in an implicit and potentially naive way. In the shrink phase, GSMN removes an edge $e_{ij}$ if at least either one of $I(X_i; X_j \mid \bm{X}_{N(i)})$ and $I(X_i; X_j \mid \bm{X}_{N(j)})$ is smaller than $\lambda$, i.e., $\min\{I(X_i; X_j \mid \bm{X}_{N(i)}), I(X_i; X_j \mid \bm{X}_{N(j)})\} \leq \lambda$, even if the other one was very large. Such treatment of symmetricity can leads to improper results especially when there are asymmetric dependencies in an MRF, that is, $I(X_i; X_j \mid \bm{X}_{N(i)})$ and $I(X_i; X_j \mid \bm{X}_{N(j)})$ are greatly different from each other. We will revisit this perspective in the experiments section.

\subsection{Incremental association Markov blanket (IAMB)}
Incremental association Markov blanket (IAMB) \citep{tsamardinos2003algorithms} is a neighborhood learning algorithm with the grow-shrink strategy. In the grow phase, one starts with an empty set and repeats adding a node $j$ that maximizes the conditional mutual information (CMI) with node $i$, given the current neighbors $N_t(i)$, i.e.,
\begin{equation}
\begin{cases}
\displaystyle j = \argmax_{j\in V\backslash N_t[i]} I(X_i; X_j\mid \bm{X}_{N_t(i)})\\
N_{t+1}(i) \leftarrow N_t(i) \cup \{j\}
\end{cases} \nonumber
\end{equation}
To avoid selecting all variables as neighbors, the optimization step will be terminated if $\max_j I(X_i; X_j\mid \bm{X}_{N_t(i)}) \leq \lambda$, where $\lambda$ is a threshold for testing conditional independence. In the shrink phase, one starts with the learned neighborhood and repeatedly removes the nodes that have less CMI than the threshold $\lambda$. A similar structure learning algorithm using conditional entropy has also been proposed with a proof of consistency under specific assumptions \citep{netrapalli2010greedy}.

IAMB is associated with the Chow-Liu algorithm in terms of the objective function. As argued in previous work \citep{brown2012conditional}, the optimization step in IAMB can be interpreted as the greedy minimization of the following objective function:
\begin{equation}
I(X_i; \bm{X}_{V\backslash N[i]} \mid \bm{X}_{N(i)}). \label{eq: min-cmi}
\end{equation}
Minimizing the CMI in Eq.(\ref{eq: min-cmi}) is equivalent to maximizing $I(X_i; \bm{X}_{N(i)})$ in accordance with the chain rule of MI
$$
I(X_i; \bm{X}_{V\backslash \{i\}}) = I(X_i; \bm{X}_{N(i)}) + I(X_i; \bm{X}_{V\backslash N[i]}\mid \bm{X}_{N(i)}).
$$
Since the left-hand-side is constant with respect to the structure, the minimum CMI term gives the maximum MI term. We noticed that summing up the CMI in Eq.(\ref{eq: min-cmi}) for all variables gives a similar optimization problem to Eq.(\ref{eq: max-mi}), though the interpretation of neighbors varies in each algorithm. In the following section, we will show that an equivalent objective function can be obtained by using pseudolikelihood.

\section{Proposed method}
\subsection{MPLE with respect to edge set}
As mentioned in the previous section, we can learn the structure of an MRF by finding the local neighborhood for each node. However, it has been unclear what objective function is maximized/minimized in some independence test techniques. Furthermore, these approaches seem to be unaware of the {\em symmetricity} in MRFs, that is, if a node $i$ is adjacent to another node $j$, then $j$ is adjacent to $i$, too. Here, we formulate MRF structure learning on the basis of MPLE with respect to the edge set $E$. The proposed reformulation leads to an optimization problem similar to that of IAMB, but the resulting solution is different from that of IAMB due to the symmetricity in MRFs.

\begin{algorithm}[t]
  \KwData{Set of random variables $\bm{X} \equiv \{X_{i\in V}\}$.}
  \Parameter{Threshold $\lambda$.}
  \KwResult{Edge set $E$.}
  \% Initialization\\
  $t \leftarrow 0$,~~ $E_0 \leftarrow \{\}$\;
  \% Grow phase\\
  \While{$\mathrm{true}$}{
    $\displaystyle i, j = \argmax_{i\in V,~ j\in V\backslash N_t[i]} \bar{I}_{ij}^G(E_t)$\;
    \If{$\bar{I}_{ij}^G(E_t) \leq \lambda$}{
      break\;
    }
    $E_{t+1} \leftarrow E_t \cup \{e_{ij}\}$\;
    $t \leftarrow t + 1$\;
  }
  \% Shrink phase\\
  \While{$\mathrm{true}$}{
    $\displaystyle i, j = \argmin_{i\in V,~ j\in N_t(i)} \bar{I}_{ij}^S(E_t)$\;
    \If{$\bar{I}_{ij}^S(E_t) > \lambda$}{
      break\;
    }
    $E_{t+1} \leftarrow E_t \backslash \{e_{ij}\}$\;
    $t \leftarrow t + 1$\;
  }
  \caption{Details of GS-MPLE.}
  \label{alg: greedy-structure-learning}
\end{algorithm}

Given a set of random variables $\bm{X}$ and a likelihood $p(\bm{x}\mid \mathcal{H}, \Theta)$, pseudolikelihood is defined as Eq.(\ref{eq: pseudo-likelihood}). Again let us define $N_E(i) \equiv \{j \mid j\neq i,~~ e_{ij} \in E ~~\text{or}~~ e_{ji} \in E\}$ and $N_E[i] \equiv N_E(i) \cup \{i\}$. To avoid restricting the range of application, we do not assume an explicit form of the underlying distribution, i.e., $(\mathcal{H}, \Theta) = (\varnothing, \varnothing)$, and consider optimization over edge set. However, pseudolikelihood is typically not a function of edge set, thus we cannot optimize the ordinary pseudolikelihood with respect to edge set. Keeping that in mind, we define {\em pseudolikelihood of edge set}
\begin{equation}
\mathcal{L}(E) \equiv \prod_{i=1}^D p(x_i \mid \bm{x}_{N_E(i)}), \label{eq: pseudo-likelihood-2}
\end{equation}
for $E\in \mathcal{E}$, where $\mathcal{E}$ denotes all possible edge sets that connects vertices in $V$. Each conditional distribution in Eq.(\ref{eq: pseudo-likelihood-2}) is an unknown but exact density/mass function without restriction. Note that Eq.(\ref{eq: pseudo-likelihood-2}) is different from ordinary pseudolikelihood; therefore, theoretical properties such as consistency of MPLE is no longer guaranteed in our formulation.

For the sake of the connection to CMI, we minimize the expected value of negative log pseudolikelihood over the true distribution $p(\bm{x})$
\begin{align}
J(E) & \equiv \langle -\ln\mathcal{L}(E) \rangle_{p(\bm{x})} \nonumber\\
& = \sum_{i=1}^D \left\langle \ln\frac{p(x_i\mid \bm{x}_{V\backslash \{i\}})}{p(x_i \mid \bm{x}_{N(i)})} \right\rangle_{p(\bm{x})} + C \nonumber\\
& = \sum_{i=1}^D \left\langle \ln\frac{p(x_i, \bm{x}_{V\backslash N[i]}\mid \bm{x}_{N(i)})}{ p(x_i \mid \bm{x}_{N(i)})p(\bm{x}_{V\backslash N[i]}\mid \bm{x}_{N(i)}) } \right\rangle_{p(\bm{x})} + C \nonumber\\
& = \sum_{i=1}^D I(X_i; \bm{X}_{V\backslash N[i]} \mid \bm{X}_{N(i)}) + C. \label{eq: npl}
\end{align}
Note that $C \equiv \sum_{i=1}^D \langle -\ln p(x_i\mid \bm{x}_{V\backslash \{i\}})\rangle_{p(\bm{x})}$ is constant with respect to $E$ and thus can be safely ignored. It can be confirmed from non-negativity of CMI that $\bm{X}$ satisfies the local Markov property when $J(E)$ is minimized. To avoid the trivial minimum that selects the complete graph, we define the optimization problem as follows:
\begin{equation}
\min_{E\in \mathcal{E}}~ \{\abs{E} \,\big\vert\, J(E) = \min_{E^\prime\in \mathcal{E}} J(E^\prime)\}.
\end{equation}
This is the smallest set of edges, such that the objective function $J(E)$ is minimized.

A trick in the proposed formulation is the expectation in $J(E)$. Since we do not know the true conditional distribution in Eq.(\ref{eq: pseudo-likelihood-2}), we cannot compute the pseudolikelihood directly from a single observation. Still, we can compute $J(E)$ from observations by calculating the CMIs. In practice, we approximate the true distribution by empirical distribution, and then the expected log pseudolikelihood equals the log pseudolikelihood of samples.

\subsection{Grow-shrink MPLE (GS-MPLE)}
To obtain the smallest edge set that minimizes the objective function, we follow the same grow-shrink strategy as GSMN and IAMB. In the grow phase, we begin with $E_0=\{\}$ and repeated add an edge to $E$. Let $J_t, E_t$, and $N_t$ denote the objective function, the selected edges, and the neighborhood at time $t$, respectively. Let us define
$$
\bar{I}_{ij}^G(E) \equiv I(X_i; X_j\mid \bm{X}_{N(i)}) + I(X_i; X_j\mid \bm{X}_{N(j)}),
$$
for $j \not\in N[i]$. Then, the following step
\begin{equation}
\begin{cases}
\displaystyle i, j = \argmax_{i\in V,~ j\in V\backslash N_t[i]} \bar{I}_{ij}^G(E_t) \\
E_{t+1} \leftarrow E_t \cup \{e_{ij}\} \label{eq: sl-forward-step}
\end{cases}
\end{equation}
will add an edge that generates the largest possible decrease in the objective function. This can be confirmed by using the following equation:
$$
J_{t+1} = J_t - \bar{I}_{ij}^G(E_t),
$$
which can be derived from the chain rule of MI. To obtain the sparsest structure, we do not add a new edge if
\begin{equation}
\max \bar{I}_{ij}^G(E_t) \leq 0. \label{eq: misl-stop}
\end{equation}

The shrink phase can be developed similarly, starting with the learned graph and removing edges iteratively. Let us define
$$
\bar{I}_{ij}^S(E) \equiv I(X_i; X_j\mid \bm{X}_{N(i)\backslash \{j\}}) + I(X_i; X_j\mid \bm{X}_{N(j)\backslash \{i\}}),
$$
for $j \in N(i)$. Then, the following step
\begin{equation}
\begin{cases}
\displaystyle i, j = \argmin_{i\in V,~ j\in N_t(i)} \bar{I}_{ij}^S(E_t) \\
E_{t+1} \leftarrow E_t \backslash \{e_{ij}\} \label{eq: sl-backward-step}
\end{cases}
\end{equation}
will remove an edge that generates the smallest possible increase in the objective function. As in the grow phase, we stop the iteration if
$$
\min \bar{I}_{ij}^S(E_t) > 0.
$$
We will refer to the proposed structure learning algorithm as {\em grow-shrink maximum pseudolikelihood estimation (GS-MPLE)} of MRFs in the rest of this paper.

A major difference between GS-MPLE and the existing algorithms is due to the fact that adding/removing an edge $e_{ij}$ will affect two CMIs, $I(X_i; \bm{X}_{V\backslash N[i]}\mid \bm{X}_{N(i)})$ and $I(X_j; \bm{X}_{V\backslash N[j]}\mid \bm{X}_{N(j)})$, in our objective function. This allows us to treat the symmetricity in MRFs explicitly whereas existing algorithms take more implicit and potentially naive way.

\subsection{Edge regularization}
Although the proposed method is designed to achieve the smallest edge set that minimizes the objective function $J(E)$, sometimes one wants to control sparsity of the learned graph for, for example, interpretability. To induce sparsity, we can add a regularization term for edges to the objective function:
\begin{equation}
J^\prime(E) \equiv \sum_{i=1}^D I(X_i; \bm{X}_{V\backslash N[i]} \mid \bm{X}_{N(i)}) + \lambda\lvert E\rvert,
\end{equation}
where $\lambda$ is a hyper parameter that controls the sparsity of the graph. Since we add just one edge in each step, we have
$$
J_{t+1}^\prime = J_t^\prime - \bar{I}_{ij}^G(E_t) + \lambda.
$$
The regularization parameter $\lambda$ does not affect which edge will be selected in each step, but it does change the stopping criterion. To avoid adding an edge that increases the objective function, we modify the threshold in Eq.(\ref{eq: misl-stop}) to
\begin{equation}
\bar{I}_{ij}^G(E_t) \leq \lambda. \label{eq: misl-stop-2}
\end{equation}
One can easily confirm that setting $\lambda = 0$ gives an optimization step without regularization. Hopefully, the thresholding in Eq.(\ref{eq: misl-stop-2}) can be interpreted as conducting a conditional independence test, though a typical conditional independence test cannot be simply applied to $\bar{I}_{ij}^G(E_t)$, the sum of two CMI terms. The details of the algorithm are summarized in Algorithm \ref{alg: greedy-structure-learning}.

\begin{figure}[t]
  \centerline{
    \subfigure{
      \includegraphics[width=0.25\textwidth]{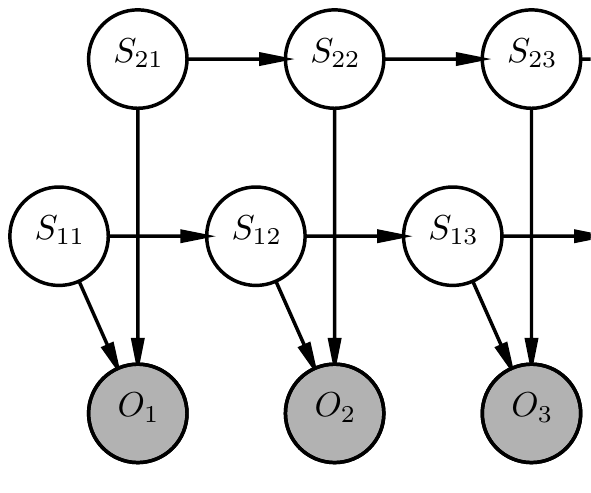}
      \label{fig: bn-hmm}
    }
    \subfigure{
      \includegraphics[width=0.25\textwidth]{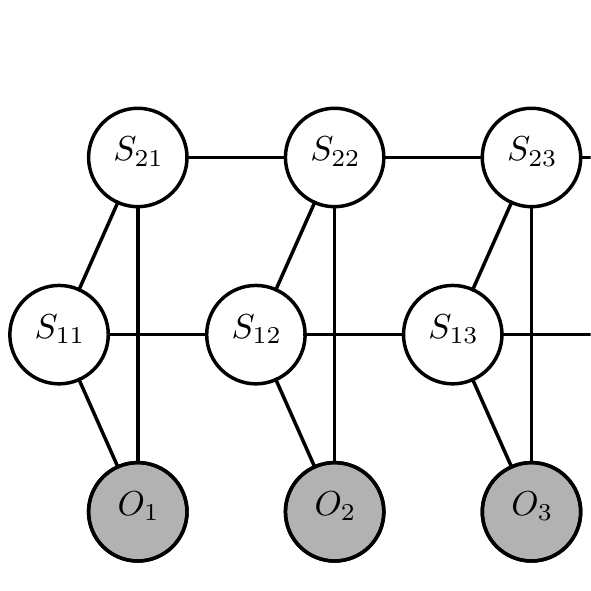}
      \label{fig: mrf-hmm}
    }
  }
  \caption{Bayesian network (left) and its moralized MRF (right) for (b) and (d).}
  \label{fig: bn-mrf-hmm}
\end{figure}

\begin{figure*}[!t]
  \centerline{
    \subfigure[Ising model]{
      \includegraphics[width=0.5\textwidth]{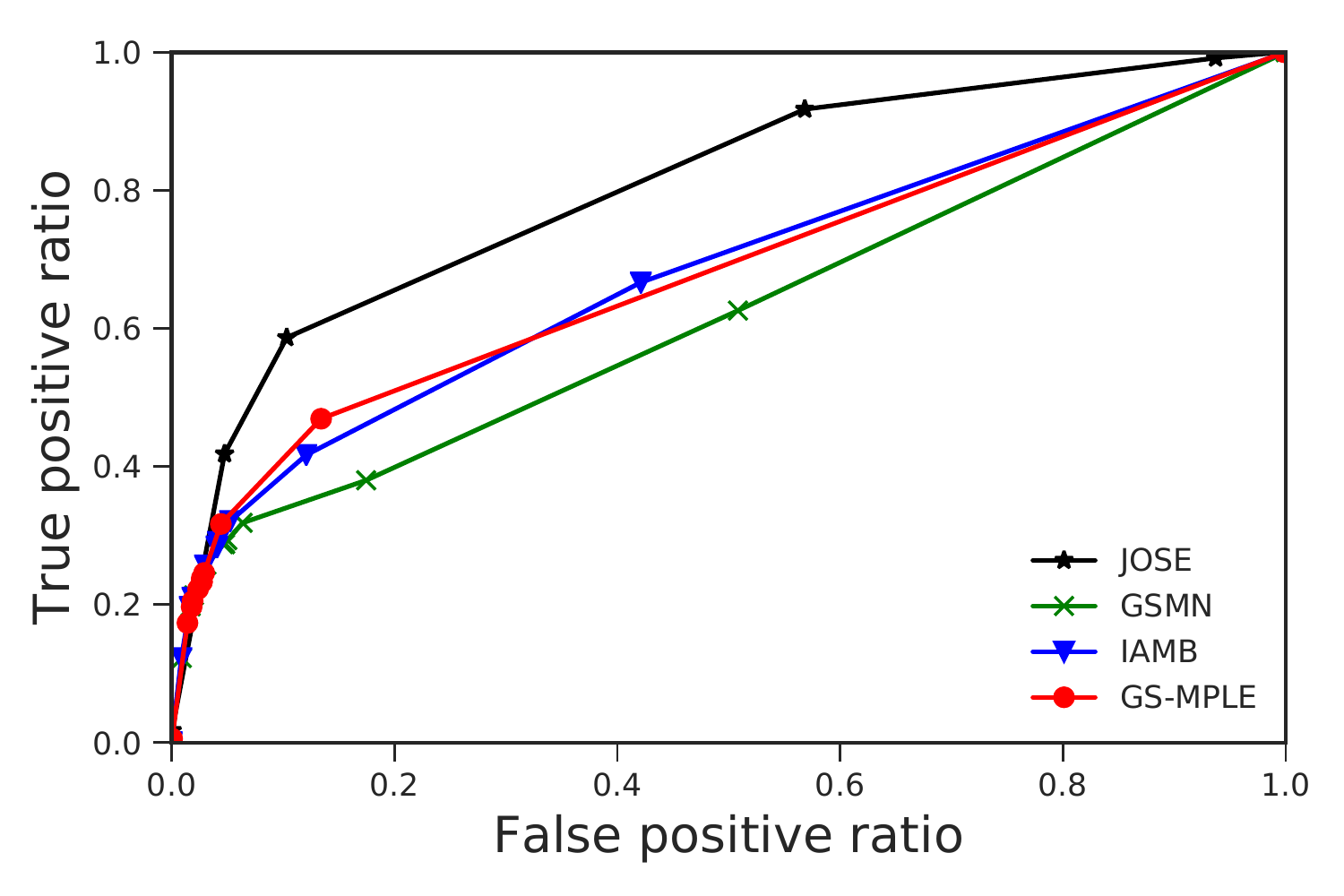}
      \label{fig: roc-ising}
    }
    \subfigure[Hidden Markov model (discrete)]{
      \includegraphics[width=0.5\textwidth]{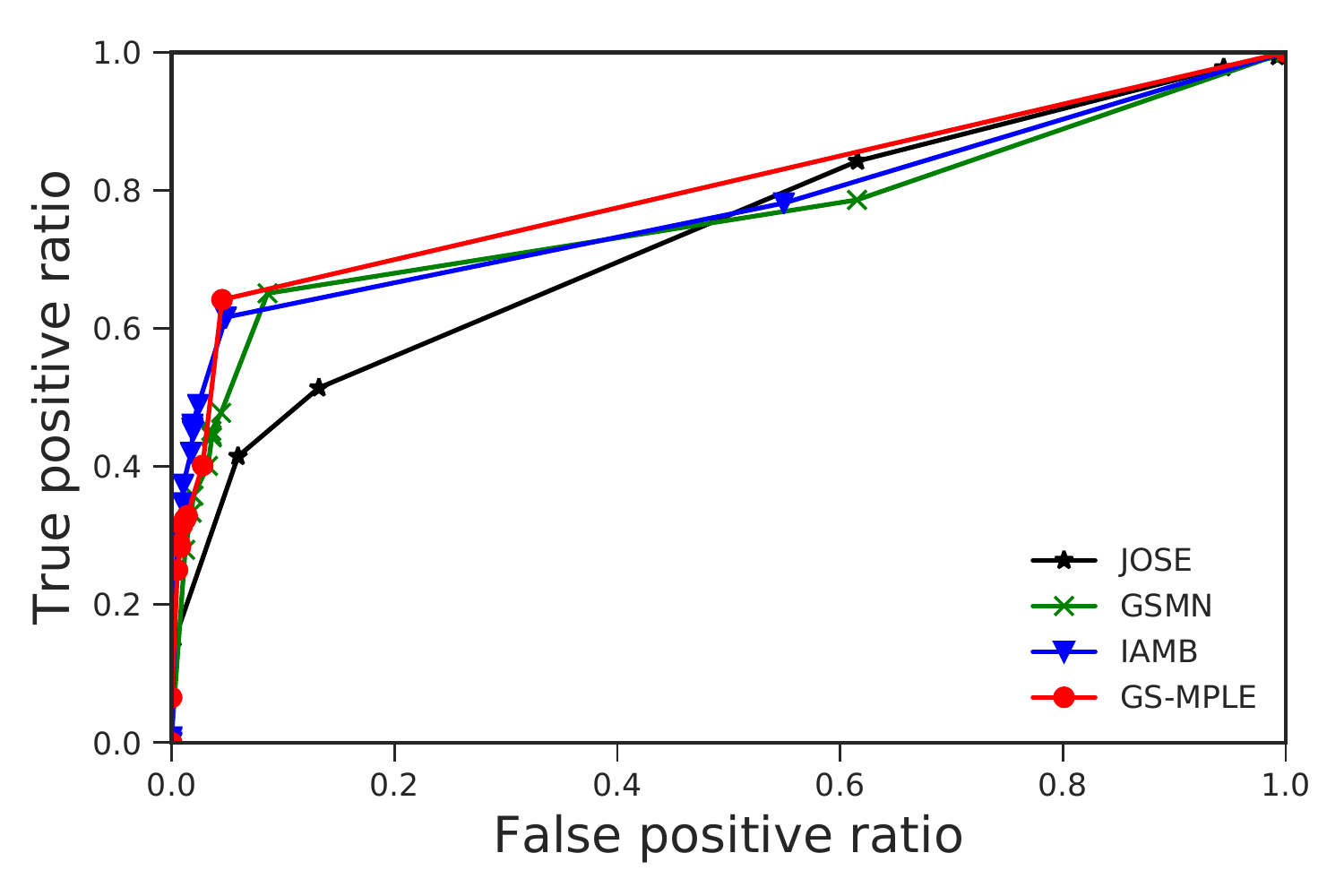}
      \label{fig: roc-hmm-d}
    }
  }
  \centerline{
    \subfigure[Guassian distribution]{
      \includegraphics[width=0.5\textwidth]{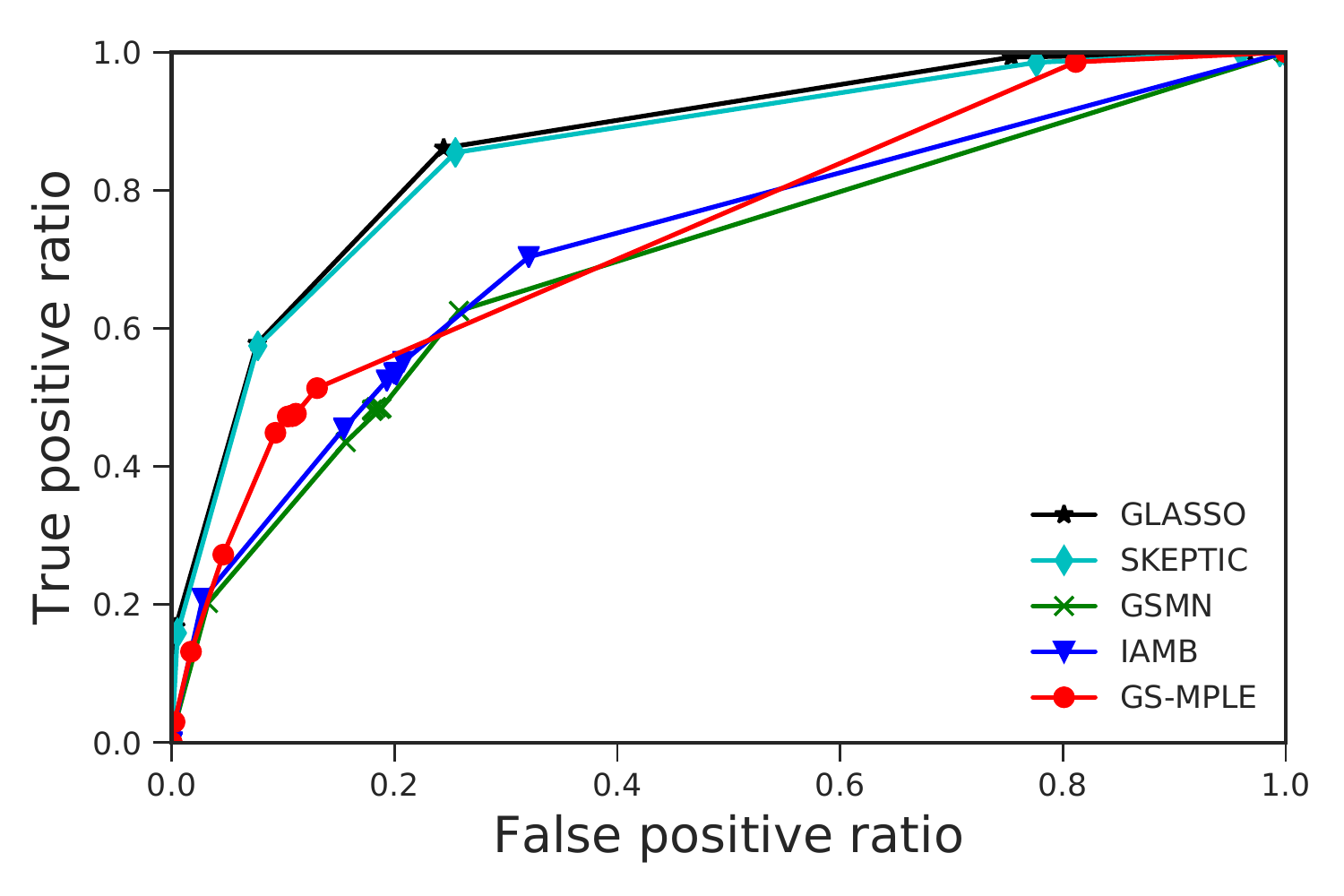}
      \label{fig: roc-gaussian}
    }
    \subfigure[Hidden Markov model (continuous)]{
      \includegraphics[width=0.5\textwidth]{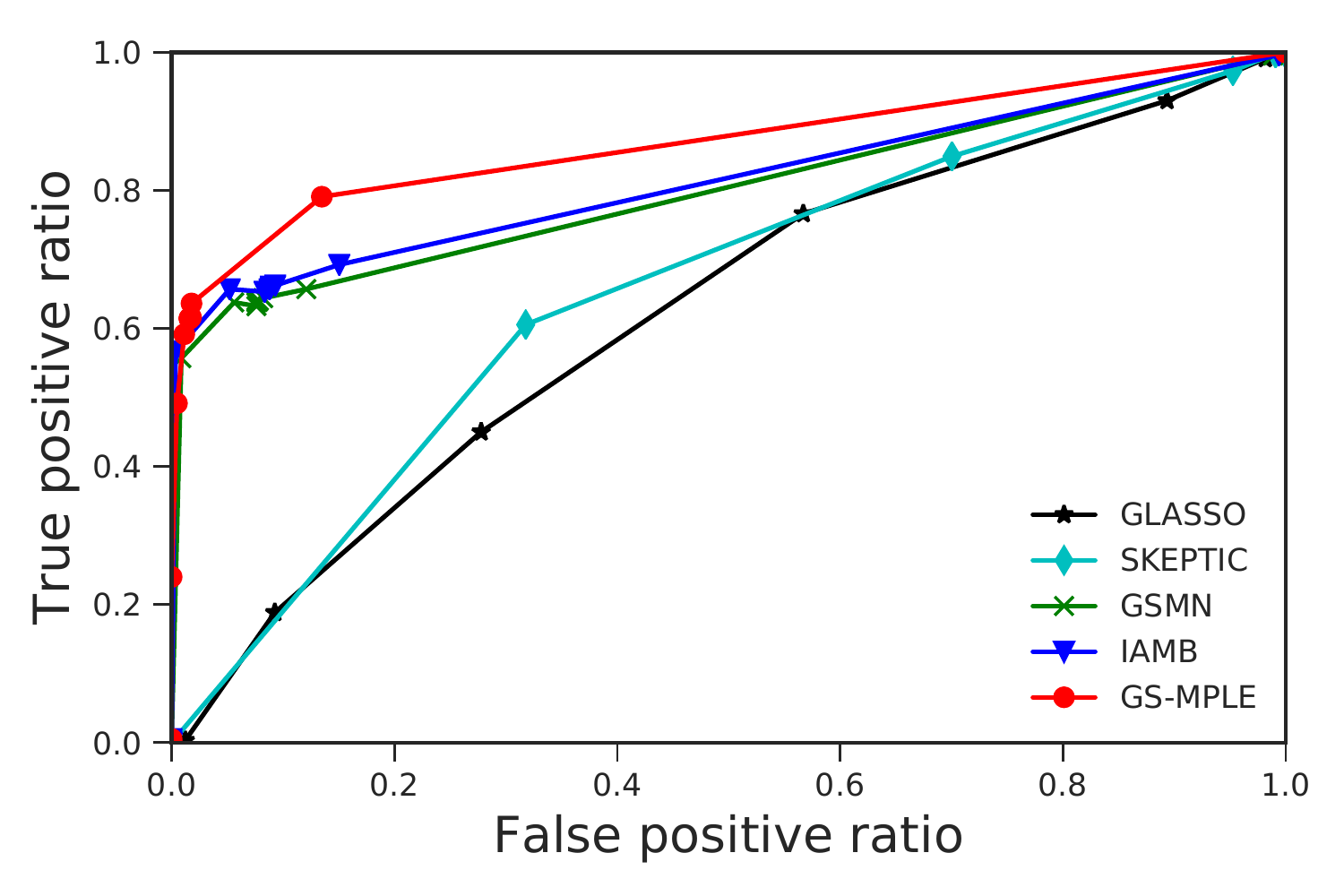}
      \label{fig: roc-hmm-c}
    }
   }
  \caption{Receiver operating characteristic (ROC) curves on edge recovery for MRFs.}
  \label{fig: roc}
\end{figure*}

\section{Experiments}
We experimentally assessed the structure learning performance of the proposed algorithm on synthetic datasets with $N=100$ observations derived from four distributions.
\begin{enumerate}[label={(\alph*)}]
\item{Square lattice Ising model with $3\times 3$ grid}
\item{Discrete hidden Markov model (HMM)}
\item{Gaussian distribution}
\item{Continuous HMM (a.k.a. state-space model)}
\end{enumerate}
We set the number of variables $D=9$ in all experiments. In addition to the MI-based structure learning algorithms (GSMN and IAMB), we used JOSE \citep{guo2010joint} as the baseline for (a) and (b) and the graphical lasso (GLASSO) \citep{friedman2008sparse} and SKEPTIC \citep{liu2012high} as the baseline for (c) and (d). We observed the true positive ratio (TPR) and false positive ratio (FPR) of edge recovery, changing hyper parameters in these methods.

To compute CMI, we used the decomposition of CMI into MIs, i.e.,
$$
\hat{I}(\bm{X}; \bm{Y}\mid \bm{Z}) = \hat{I}(\bm{X}, \bm{Z}; \bm{Y}) - \hat{I}(\bm{Y}; \bm{Z}),
$$
where $\bm{X}$, $\bm{Y}$, and $\bm{Z}$ are random variables, and $\hat{I}(X; Y)$ is the MI estimated by the extended KSG estimator for discrete-continuous mixtures \citep{gao2017estimating}.

For (b) and (d), we consider a variant of HMM that has two variables as the hidden state at each time. Each state transitions separately, and the observed variable depends on both hidden states. Figure \ref{fig: bn-mrf-hmm} shows the Bayesian network and its moralized MRF for the model. The intuition behind using this model is to ensure the data contain {\em ternary} relationships, meaning that conditioning by a variable {\em increases} information shared by the other two variables. Each variable in (b) the discrete HMM takes binary values, and transition and observation are defined by conditional probability tables (CPTs). In (d) the continuous HMM, transition is defined as a linear transformation with Gaussian noise, while the observed variable at time $t$ is derived from the Gaussian distribution $\mathcal{N}(s_{1t}, s_{2t}^2)$, where $s_{1t}$ and $s_{2t}$ denote the first and second state variables at time $t$.

Figure \ref{fig: roc} shows the receiver operating characteristic (ROC) curves of the average TPR and FPR of $100$ independent runs. The parameter-learning based approaches, such as JOSE, GLASSO and SKEPTIC, outperformed MI-based ones in (a) and (c) in which the assumptions of these methods are satisfied. In contrast, these parametric models performed poorly in (b) and (d), which have {\em ternary} relationships caused by the head-to-head structures in the model. MI-based approaches tend to be more robust to the {\em ternary} relationships than parameter-learning based ones.

Moreover, it is worth noting that GS-MPLE outperforms the other independence-test-based methods in (d). This is possibly explained by the asymmetric dependencies in (d). In the continuous HMM, we derived the observed variable $O_t$ from the Gaussian distribution with mean $S_{1t}$ and standard deviation $S_{2t}$. Consequently, the degree of influence from each state to the observed variable can differ significantly (that is, the two CMIs in $\bar{I}_{ij}^G(E)$ can be greatly different from each other), whereas the discrete HMM treats the hidden states in the same way.

\section{Discussions}
Though we conducted experiments only in purely discrete and purely continuous settings, we can apply GS-MPLE and other independence-test-based approaches to discrete-continuous mixed setting, using the MI estimator for discrete-continuous mixtures \citep{gao2017estimating}. In the future, structure learning performance in mixed settings should be examined.

It might be noted that the worst-case time complexity of GS-MPLE is $O(D^4 K)$, where $D$ is the number of random variables and $K$ is the worst-case time complexity of the CMI estimator. This can be confirmed as follows. In the worst case, the {\em while} loop runs $O(D^2)$ times. For every iteration of the {\em while} loop, CMIs are calculated at most $O(D^2)$ times. Thus, the overall time complexity is $O(D^4 K)$. $K$ depends on what CMI estimator we use. Specifically, in the purely discrete case, a naive CMI decomposition to entropies gives the worst-case time complexity $O(ND)$, where $N$ is the sample size. Then, the overall time complexity is $O(ND^5)$. This is in fact slower than the worst-case time complexity $O(ND^4)$ of the existing neighborhood-selection-based algorithms \citep{tsamardinos2003algorithms, netrapalli2010greedy}. Thus, there seems to be a trade-off between accuracy and computational efficiency.

We also note that, with the rise of statistical relational models in recent years, combining probabilistic graphical models and first order logic has gained more attention. While this paper focuses on ordinary graphical models, the result will (hopefully) serve as a basis for learning structures of statistical relational models in the future.

\section{Conclusions}
We described a novel structure learning algorithm for Markov random fields (MRFs) based on maximum pseudolikelihood estimation with respect to the edge set. The proposed solution differs from the existing independence-test-based algorithms in that it explicitly takes into account symmetricity in MRFs. Experiments showed that the proposed structure learning algorithm performs as well as, or sometimes better than, the existing methods based on independence test. In the future, research on theoretical properties such as consistency of the proposed formulation is needed as well as more rigorous experiments on larger datasets. In addition, it is also desired to apply the proposed algorithm to learning structures of statistical relational models.

\section*{Acknowledgement}
This work was supported by JSPS Grant-in-Aid for Scientific Research on Innovative Areas (Grant Number JP25120009) and Cybozu Labs youth.

\bibliographystyle{aaai}
\bibliography{ijcaiws2018}

\end{document}